\documentclass[a4paper,11pt]{article}
\usepackage{url}
\usepackage{amsmath}
\usepackage{amsfonts}
\usepackage{amssymb}

\title{Text Classification with Compression Algorithms}
\author{Antonio G. Zippo\\
}

\date{}

\begin{document}
\maketitle

\begin{abstract}
This work concerns a comparison of SVM kernel methods in text categorization tasks. In particular I define a kernel function that estimates the similarity between two objects computing by their compressed lengths. In fact, compression algorithms can detect arbitrarily long dependencies within the text strings. Data text vectorization looses information in feature extractions and is highly sensitive by textual language. Furthermore, these methods are language independent and require no text preprocessing. Moreover, the accuracy computed on the datasets (Web-KB, 20ng and Reuters-21578), in some case, is greater than Gaussian, linear and polynomial kernels. The method limits are represented by computational time complexity of the Gram matrix and by very poor performance on non-textual datasets.
\end{abstract}

\section{Introduction}
In the world of discrete sequences (or sequential data), learning problem is an important challenge in pattern recognition and machine learning. Classification tasks that involve symbolic data are very frequent. For instance, text categorization tasks, e.g. news, web pages and document classification, are widely employed. In such tasks, classification algorithms like support vector machine, neural networks and many others require the conversion of these symbolic sequences into feature vectors \cite{compml}. This preprocessing typically looses information. For instance, stemming phase maps words like {\it showing}, {\it shows}, {\it shown} into the same representative (suffix-free) feature word {\it show}. Furthermore, this stage is very language-dependent and sensitive, i.e. an english text stemmer is very different from a spanish or a russian text stemmer. Finally, other preprocessing procedures remove stop and short words.\\ 
I employed a novel framework based on a different perspective. Textual data are treated as symbol sequences and by mining the structure of these sequences it is possible to define a similarity measure between sequence pairs. That's  the definition of a kernel function over the features space. Thus a learning phase is needed to capture features from the given sequences. Then a similarity measure is required to quantify the shared features in sequence couples. Finally the kernel trick allows for applying classification algorithms like SVMs. The aim of this work is to compare results obtained from classical kernels with a compression based similarity kernel, i.e. the Normalized Compression Distance (NCD) \cite{phdthesis,cilibrasi,infodist,similarmetric}. Both methods were exerted on Web-KB \cite{webkb}, two istance of Reuters-21578 \cite{davidLewis} and on 20 Newsgroups \cite{20ng} datasets.\\
The section \ref{Methods} faces the statistical foundations of Variable Order Markov Models (VOMMs) and subsequently the definition of the $K_{NCD}$ kernel and the multiclass classification problem. In the section \ref{Experiments}, I will present the vectorization method involved in the preprocessing phase of the datasets. In the same section, I will show the results obtained for both methods and datasets. Finally a brief dissertation about the symbolic learning is faced in section \ref{Discussions}.

\section{Methods}\label{Methods}
From now, for the rest of document, I assume that $S_m = \{(x_1,y_1),\dots,(x_m,y_m)\}$ is the training set where each $x_i \in \mathbb{R}^p$ and $y_i \in \{1,\dots,M\}$. $M$ is the number of classes and $p$ is the dimensionality of training vectors. As I introduced in the previous section, text classification requires special attention on how the data and its features are represented. In addition text categorization requires an {\it ad hoc} implementation for each natural language where it is applied. A general technique able to measure the similarity between same language texts saves a lot of implementation time. Starting by presenting multiclass extension of Support Vector Machine algorithm, then I introduce the Variable Order Markov Models (VOMMs) \cite{opuvomm} underlying component of widely used compression algorithms like Prediction by Partial Matching (PPM) \cite{ppm}, Context-Tree Weighting (CTW) \cite{ctw} Lempel-Ziv Markov Chain (LZMA) \cite{lempelZiv} and Probabilistic Suffix Tree (PST) \cite{pst}
. Finally, a measure similarity function will be shown and the kernel based on this similarity measure will be presented.

\subsection{Support Vector Machine}
Support Vector Machine algorithm is a binary linear classifier that produces a separation hyperplane (whenever the training set is linearly separable) that partitions the trainining space into two classes \cite{scholkopf}. The hyperplane equation represents the decision function for all the unseen data points. On the one hand, text categorization requires, usually, more than two classes. On the other hand, linear separation is a rare condition in real-world problems. For this reason I firstly introduced a SVM algorithm able to control with special variables (slack) the non-separability of a training set. This SVM extension is called \textit{soft margin classifier}. Furthermore another important technique allows complex datasets to be linearly separable. Kernel functions, in fact, map the original dataset into a higher dimensional space where the linear separation might be done. The combination of SVM and Kernel functions becomes a very powerful technique to face very complex classification problems.\\
First, I present the quadratic optimization problem in dual form without slack variable:
\begin{equation}\label{eq:originalDual}
\begin{aligned}
& \underset{\alpha \in \mathbb{R}^m}{\text{maximize}}
& & \sum_{i=1}^m \alpha_i - \frac{1}{2}\sum_{i,j=1}^m \alpha_i\alpha_j y_i y_j \left<x_i,x_j\right> \\
& \text{subject to}
& & \alpha_i \geq 0,\; i=1,\dots,m\;\\
& & &\sum_{i=1}^m \alpha_i y_i = 0.
\end{aligned}
\end{equation}
where $\alpha_i$ are the Lagrangian multiplier inherited from primal to dual problem conversion and $\left<x,y\right>$ is the inner product (within the inner product space) between $x$ and $y$. Once the optimization is done, the set of $\alpha_i$ allows for classifying a new data point $x$ with the decision function defined as
\begin{equation}\label{eq:origianalDecision}
f(x) = \text{sign} \left(\sum_{i=1}^m \alpha_i y_i \left<x,x_i\right> + b\right)
\end{equation}
Let $K(\cdot,\cdot)$ be a positive semi-definite kernel function. Thanks to Mercer's theorem, $K$ could be expressed as dot product in a higher dimensional space, i.e. $K(x,y) = \left<\phi(x),\phi(y)\right>$. The kernel trick method provides that SVM, for instance, can be combined with the kernel function to obtain a linear classification in a higher dimensional space, defined implicitly by the kernel function. The original dual problem \ref{eq:originalDual} could be rewritten as
\begin{equation}\label{eq:kernelDual}
\begin{aligned}
& \underset{\alpha \in \mathbb{R}^m}{\text{maximize}}
& & \sum_{i=1}^m \alpha_i - \frac{1}{2}\sum_{i,j=1}^m \alpha_i\alpha_j y_i y_j K(x_i,x_j) \\
& \text{subject to}
& & \alpha_i \geq 0,\; i=1,\dots,m\;\\
& & &\sum_{i=1}^m \alpha_i y_i = 0.
\end{aligned}
\end{equation}
and decision function \ref{eq:origianalDecision} becomes
\begin{equation}\label{eq:kernelDecision}
f(x) = \text{sign} \left(\sum_{i=1}^m \alpha_i y_i K(x,x_i) + b\right)
\end{equation}
Even when in feature kernel space the data points are non linearly separable, an extension of the previous problem is needed. In this case, the slack variables $\xi_i \geq 0$ constitute the relaxation of the primal form constraints $y_i(\left<x_i,w\right> + b) \geq 1-\xi_i$, with $i=1,\dots,m$. Thus, the problem \ref{eq:kernelDual} becomes:
\begin{equation}\label{softMargin}
\begin{aligned}
& \underset{\alpha \in \mathbb{R}^m}{\text{maximize}}
& & \sum_{i=1}^m \alpha_i - \frac{1}{2}\sum_{i,j=1}^m \alpha_i\alpha_j y_i y_j K(x_i,x_j) \\
& \text{subject to}
& & \alpha_i \geq 0,\; \forall i=1,\dots,m\; \\
& & &\sum_{i=1}^m \alpha_i y_i = 0\\
& & &0 \leq \alpha_i \leq \frac{C}{m} \;  i=1,\dots,m.
\end{aligned}
\end{equation}
Only the last constraint, that limits the Lagrangian multiplier values, distinguishes problem \ref{eq:kernelDual} from \ref{softMargin}.

\subsubsection{Multiclass SVM}
The natural extension of SVM binary classification problem into multiclass classification could be represented by the following optimization problem in primal form
\begin{equation*}
\begin{aligned}
& \underset{w_r \in \mathcal{H}, \xi^r \in \mathbb{R}^m, b_r \in \mathbb{R}}{\text{minimize}}
& &\frac{1}{2} \sum_{r=1}^M ||w_r||^2 + \frac{C}{m} \sum_{i=1}^m \sum_{r \neq y_i} \xi_i^r\\
& \text{subject to}
& &\left< w_{y_i},x_i \right> + b_{y_i} \geq \left< w_{r},x_i \right> + b_r + 2 - \xi_i^r\\
& & &\xi_i^r \geq 0.
\end{aligned}
\end{equation*}where $m \in \{1,\dots,M\}\backslash y_i$, $y_i \in \{1,\dots,M\}$ is the multiclass label of the pattern $x_i$. Computational issues suggest different multilabel classification strategies based on a combination of several binary classifiers. 

\paragraph{One-vs-One.}
In a first strategy, it is possible to train $M(M-1)/2$ binary classifiers for each class couples. The final decision function evaluates the decision function of every classifier and classifies the object assigning the class that obtains the highest number of votes. In this strategy, the number of binary classifier is quadratic but the computational time required for single classifier training is restricted because the data points evaluated are the number of examples belonging to the two trained classes.

\paragraph{One-vs-the Rest.}
It is otherwise possible to train $M$ binary classifiers one for every classes. In this case each classifier is trained to discriminate a class by all other classes. In this case, the decision function is defined as
\begin{equation*}
\begin{aligned}
& \underset{j \in M}{\text{argmax}}
& &\sum_{i=1}^m y_i \alpha_i^j K(x,x_i) + b^j\\
\end{aligned}
\end{equation*}
In this strategy, the number of binary classifiers are linear but the computational time required for single classifier training is higher than that used in the previous method since the number of data points evaluated is the entire training set $S_m$. Moreover, in each training stage the binary classifier is usually trained on many more negative than positive examples.\\
Notwithstanding no significative accuracy differences exist, in general, among the three methods, the \textit{one-vs-one} is used in many SVM multiclass implementations like \texttt{libsvm}.

\subsection{Variable Order Markov Models}
Data text vectorization looses information in feature extractions and it is highly sensitive to text language. In order to overcome these limitations, it is advantageous to employ sequential learning techniques that extract similarities directly on the textual learnt structure.\\ Sequential data learning usually involves quite simple methods, like Hidden Markov Models (HMM), that are able to model complex symbolic sequences assuming hidden states that control the system dynamics. However, HMM training suffers from local optima and their accuracy performance has been overcome by VOMMs. Other techniques like $N$-gram models (or $N$ order Markov Chains) compute the frequency of each $N$ long subsequence. In this case the number of possible model states grows exponentially with $N$. Both computational space and time issues arise.\\
In this perspective, the textual training sequence is generated by a stationary unknown symbol source $S=\left< \Sigma,P \right>$ where $\Sigma$ is the symbol alphabet and $P$ is the symbols probability distribution. A VOMM, given the maximum order $D$ of conditional dependencies and a training sequence $s$ generated by $S$, returns a model for the source $S$ that's an estimation $\hat{P}$ of probability distribution $P$. Applying VOMMs, instead of $N$-gram models, takes several advantages. A VOMM estimation algorithm builds efficiently a model for $S$. In fact, only the occurred $D$-grams are stored and their conditional probabilities $p(\sigma|s)\;,\sigma \in \Sigma\; \text{and } s\in \Sigma^{d \leq D}$ are estimated. This trick saves a lot of memory and computational time and makes feasible to model sequences with very long dependencies ($D \in [1,10^3]$) on 4GB personal computers.

\subsection{Lossless Compression Algorithms}
Lossless Compression Algorithms (LCAs) build a prefix tree to estimate the symbol probability distribution $P$ by combining conditional probability of a symbol with a chain rule, given $d$ previous symbols (usually $d \leq D$). In other words, LCAs produce a VOMM by some estimation algorithm in the first stage. In the second stage, LCAs compress actually the sequence applying some encoding scheme like Arithmetic Encoding (AE). The AE assigns a real value number within interval $[0,1)$ to the original sequence starting by the estimated conditional probabilities $p(\sigma|s)$, $\sigma \in \Sigma$ and $s \in \Sigma^{d \leq D}$ \cite{opuvomm}. 
Let $C(\cdot)$ be the function that computes the compressed sequence length through some GPL compressor like \texttt{bzip2},\texttt{ ppmc},\texttt{ lzma}. It is possible to prove that using average log-loss as estimation of prediction accuracy, prediction accuracy and compression ratio are equivalent \cite{rissanen}. Thus better predictions mean better compressions. Sequences easy to compress are sequences easy to learn and predict. 

\subsection{Similarity Measure}
The function $C$ brings toward the definition of a similarity measure. Once that the schemes from a sequence are detected then it is possible to measure how many of them are shared by another sequence schemes. With this aim, Cilibrasi et al \cite{infodist,similarmetric} define a similarity measure that quantify the compression facility of a sequence $x$ given the compression scheme of sequence $y$. The Normalized Compression Distance (NCD) is defined as follows:
\begin{equation}
NCD(x,y) = \frac{C(xy) - \min\{C(x),C(y)\}}{\max\{C(x),C(y)\}}
\end{equation}
where $xy$ represents the concatenation of sequence $x$ with sequence $y$ and $C(x)$ is a function that returns the length of the compressed version of $x$. The range of $NCD(x,y)$ is $[0,1]$. The $NCD(x,y) = 0$ shows that $x$ and $y$ are identical whereas $NCD(x,y) = 1$ indicates that two objects are very dissimilar. The $NCD$ function cannot work directly as a kernel function. In fact the $NCD$ function is not symmetric. The symmetry property holds defining the kernel function $K_{NCD}(x,y)$ \cite{cilibrasi} as
\begin{equation}
K_{NCD}(x,y) = 1-\frac{NCD(x,y)+NCD(y,x)}{2}
\end{equation}
However, as in many string kernels, the semidefinite positive property cannot be proved.

\section{Experiments}\label{Experiments}
I used four different datasets to test accuracy and robustness of proposed methods in comparison to other standard kernels. These datasets are collated by Ana Cardoso-Cachopo\cite{dataset}. The author split each dataset obtaining randomly two thirds of the documents for training and the remaining third for testing. The first dataset is the Web Knowledge Database (Web-KB), a collection of web pages by Carnegie Mellon University manually classified by text learning group. The second and third dataset are obtained from the Reuters-$21578$ dataset, that's the collection of classified Reuters news restricted to eight classes (R8) or fiftytwo (R52). Finally, the last dataset represents a collection of approximately $20000$ newsgroup documents collected by Ken Lang. The Table \ref{datasetFeatures} reports the number of classes, number of training documents, number of testing documents and number of features for each dataset. For futher details consult the web page \cite{dataset}.\\ 
Every dataset has been processed following four stages:
\begin{enumerate}
 \item Terms are extracted from document. All letters are converted into lowercase and trimming of tabulations, multispaces and non-visible characters are done.
 \item Removing of less-than-3-characters-long terms
 \item Stopwords removing 
 \item Applying a stemming procedure
\end{enumerate}
For experiments with classical kernels, I used the $4^{th}$ stage stemmed datasets in order to decrease as many as possible features. The final vectorized training/testing set represents the count of each appeared terms. Before, the training/testing stage, the dataset feature vectors are scaled into $[-1,1]$ to prevent overfitting. The whole experimental stages are performed using Python programming language and I investigated the accuracy of the proposed kernel with the {\tt scikits.learns} Python package that it's bound to the \texttt{libsvm} SVM implementation. Results that appear in Table \ref{tableResults} represent the best accuracy on the test datasets after a cross-validation procedure for the choice of the best model. To employ the $K_{NCD}$ kernel, I used the first stage datasets to compute the Gram matrix $G = [k(x_i,x_j)],\;$ with $i,j=1,\dots,m$. Although, the Gram matrix computation waste a lot of computational time, the {\tt complearn-tools} package included in all Debian based Linux 
distributions (like Ubuntu) requires at most $5$-$20$ minutes to compute the matrix thanks to its efficient multicore implementation \cite{complearn}. The experiments ran on a Dell Precision workstation with 24 GB Ram and dual Quadcore Xeon X5677 at 3.46 Ghz.\\
Furthermore, $K_{NCD}$ SVM were employed to perfom another pratical classification task as handwritten recognition. In this case I used the $0$-$9$ digit MNIST dataset. Results from the unsatisfactory experiment are not reported due to very disastrous performaces. The overall accuracy never exceeds $54.2\%$. A discussion about this failure is reported in Section \ref{Discussions}.\\
The accuracy of $K_{NCD}$ SVM kernel is higher, in some case, than that achieved by the classical SVM kernels like Gaussian, polynomial and linear. The results are shown in the Table \ref{tableResults}. The showed results are obtained after $K$-fold cross-validation (with $K=5$) sessions to fit the best kernel parameters that are reported in Table \ref{tableParameters}. For $K_{NCD}$ and linear kernels C is the only reasonable parameter that can influence the accuracy rather than the polynomial and Gaussian kernels that have other important parameters like $d$,$\gamma$ and $r$. The model selection procedure computes the mean accuracy of model with five-fold cross-validation and then stores the obtained result. Once that the procedure tests every admissible values for each parameter, the parameter combinations with higher accuracy is returned.
\begin{table}
\begin{center}
    \begin{tabular}{ | l | l | l | l | l |}
    \hline
    Dataset & \#Classes & \#Train Docs& \#Test Docs & \#Features \\ \hline
    Web-KB & 4 & 2803 & 1396 & 7770 \\ \hline
    R8 & 8 & 5485 & 2189 & 17387 \\ \hline
    R52 & 52 & 6532 & 2568 & 19241 \\ \hline
    20ng & 20 & 11293 & 7528 & 70216 \\ \hline
    \end{tabular}
\end{center}
\caption{Dataset characteristics. The column values represent respectively the number of classes, the dimension of training set, the dimension of testing set and the number of features.}\label{datasetFeatures}
\end{table}
\begin{table}
\begin{center}
    \begin{tabular}{ | l | l | l | l | l |}
    \hline
    Dataset & $K_{NCD}$ & Linear& Polynomial & Gaussian \\ \hline
    Web-KB & 94.38\% & 85.82\% & 94.11\% & 50.87\% \\ \hline
    R8 & 94.33\% & 96.98\% & 94.42\% & 49.67\% \\ \hline
    R52 & 89.48\% & 92.39\% & 90.00\% & 49.82\% \\ \hline
    20ng & 87.71\% & 84.26\% & 86.81\% & 48.27\% \\ \hline
    \end{tabular}
\end{center}
\caption{Accuracy of the proposed kernels on the four testing sets.}\label{tableResults}
\end{table}
\begin{table}
\begin{center}
    \begin{tabular}{ | l | l | l | l | l |}
    \hline
    Dataset & $K_{NCD}$ & Linear& Polynomial & Gaussian \\ \hline
    Web-KB & C=4 & C=0.07 & C=0.1,$d=6$,$\gamma=0$,$r=2$ & C=7,$\gamma=9$,$d=4$ \\ \hline
    R8 & C=3 & C=1.5 & C=0.1,$d=7$,$\gamma=0.1$,$r=2$ & C=0.8,$\gamma=3$,$d=2$ \\ \hline
    R52 & C=1 & C=2.8 & C=0.1,$d=7$,$\gamma=0.1$,$r=2$ & C=1.4,$\gamma=2$,$d=2$ \\ \hline
    20ng & C=11 & C=0.01 & C=2.3,$d=6$,$\gamma=0.1$,$r=2$ & C=5,$\gamma=0$,$d=1$ \\ \hline
    \end{tabular}
\end{center}
\caption{Chosen SVM and kernel parameters after  $K$-fold cross-validation with $K=5$ over the training sets. The admissible values are respectively $\{0.01,0.02,\dots,0.1,0.2,\dots,3,4,\dots,20\}$ for C, $\{1,2,\dots,19\}$ for $d$, $\{0,0.1,\dots,1\}$ for $\gamma$ and $\{0,1,\dots,6\}$ for $r$}\label{tableParameters}
\end{table}

\section{Discussion}\label{Discussions}
Kolmogorov complexity $K$ of an object $x$ expressed as string (or symbol sequence) represents the length of the shortest program, for a universal Turing machine, that outputs the $x$ string. In other words, the Kolmogorov complexity, measures the amount of useful knowledge to compute a given object that is the semantic object content. The Kolmogorov Complexity is uncomputable and this can be proved by the reduction from the uncomputability of the Halting Problem. The first important inequality is that:
\begin{equation*}
K(x) \leq |x| + c, \forall x
\end{equation*}
where $c$ is a costant and $|x|$ is the $x$ length.\\ Some information contents are syntactically accessible, some others not. For instance, considering the digits of the natural constant $\pi$, no syntactic information can be extracted. In fact $\pi$ (as many other natural constant) passes every randomness test. No structure can be extracted only from it's digits. Nevertheless it is quite simple to write a short computer program that outputs the $\pi$ digits. Thus, only semantic information allows a $\pi$ digits compression. However many symbolic sequences involved in real-world problems could be syntatically compressed. Moreover, many symbolic schemes are unaccessible by a human observer because the obvious undetectability of million symbol long recurrences within a string.\\
Lossless compression algorithms allow syntactic compression of an object like a binary string. The basic idea is that given a fixed object, a compression algorithm is able to rewrite the object such that the length of the rewritten version is smaller than the original version length. The reduced object length proves the compression algorithm capacity to describe the object in terms of rules and schemes. Hence the compression algorithm abilities purely act on a syntactic level. In this way, the compressor code imposes an upper bound to Kolmogorov complexity. This upper bound is stronger than the previous inequality since:
\begin{equation*}
K(x) \leq C(x) \leq |x| + O(1), \forall x
\end{equation*}
where $C(x)$ is the $x$ compressed version length. The idea that compressor codes could approximate Kolmogorov complexity was first presented in some works \cite{phdthesis,cilibrasi,infodist} that brought to the definition of a similarity metric called Normalized Compression Distance and of a kernel based on it. Their results showed successful applications with unsupervised and supervised tasks such as text categorization, protein and music clustering.\\
Learning of sequential data remains still an open challenge. VOMMs obtain good results in several classification tasks on symbolic data \cite{opuvomm}. I remark as the NCD function is a feature-free distance function, i.e. the similarity estimation it is not based on some fixed features. On the contrary every other similarity measure is feature-based, i.e. requires detailed knowledge of the problem area in order to measure the similarity/dissimilarity between two objects.\\
The failure of $K_{NCD}$ kernel on numerical datasets can be understood by an example. Considering the sequences $s =``1.999999``$ and $r=``2.000000``$, their meanings (the quantities) are very close while the symbolic sequences are very dissimilar having no common symbols. The same thing happens with synonyms in textual data. Again $grow$ and $arise$ are considered very dissimilar. However, in real-world problems, situation like the latter are rare, while former ones are very common. In addition for a given object, the number of potential neighbors is an order of magnitude greater for numerical objects than for textual objects. 


\section{Conclusions}
The accuracy of proposed kernel outperforms the accuracy of standard kernels with some datasets (20ng and Web-KB). The $K_{NCD}$ kernel method cannot carry out a classification task in general. In fact, compression based methods fail on numerical dataset because numbers (and their digits) enclose a coding, e.g. integer numbers. Furthermore computational time complexity constitutes a feasibility problem for a lot of pratical tasks. Nevertheless the good results highlight this promising framework. The $K_{NCD}$ kernel is independent by document languages and could be utilized with eastern ideographic languages.

\end{document}